\documentclass{article}
\pdfoutput=1
\usepackage{acra}
\usepackage{graphicx} % for pdf, bitmapped graphics files
\usepackage{siunitx}
\usepackage{amsmath} % assumes amsmath package installed
\usepackage{amssymb}  % assumes amsmath package installed
\usepackage{enumitem}
\usepackage{float}
\usepackage{xcolor}
\usepackage[caption=false, font=footnotesize]{subfig} 
\usepackage{titlesec}
\usepackage{gensymb}
\usepackage{tabularx,booktabs}
\usepackage[font=footnotesize]{caption}
\usepackage{adjustbox}
\newcommand{\w}{4}
\newcommand{\h}{2.5}

\newcommand{\hIII}{4.5}

\usepackage{todonotes}

\usepackage{babel}
\newcolumntype{x}[1]{>{\centering\let\newline\\\arraybackslash\hspace{0pt}}m{#1}}
\title{Guided Curriculum Learning for Walking Over Complex Terrain}
\author{Brendan Tidd \\ Queensland University of Technology \\ brendan.tidd@hdr.qut.edu.au \And Nicolas Hudson \\ CSIRO \\ nicolas.hudson@csiro.au \And Akansel Cosgun \\Monash University \\ akansel.cosgun@monash.edu}

\begin{document}

\maketitle

%%%%%%%%%%%%%%%%%%%%%%%%%%%%%%%%%%%%%%%%%%%%%%%%%%%%%%%%%%%%%%%%%%%%%%%%%%%%%%%%
\begin{abstract}
%Reliable bipedal walking over complex terrain is a challenging problem. Curriculum learning mimics how humans learn, first starting with an achievable version of a task, and increasing the difficulty as milestones are met. In this work, we apply a curriculum learning approach to biped walking and show that our method can successfully traverse complex terrain artifacts, where without the curriculum the agent takes much longer to learn, or in some cases fails to learn. For each terrain type (flat, hurdles, gaps, stairs, steps, and jumps), we train a separate DRL policy by applying a curriculum of three stages. We first train a policy with expert guided forces, incrementing the difficulty of the terrain. We then decay the guiding forces to zero, before finally increasing perturbations applied to the agent to improve robustness. We demonstrate that a simple hand designed walking trajectory is a sufficient prior to improve learning outcomes for a suite of complex terrain types, where we do not have access to specific demonstrations for all terrains.

Reliable bipedal walking over complex terrain is a challenging problem, using a curriculum can help learning. Curriculum learning is the idea of starting with an achievable version of a task and increasing the difficulty as a success criteria is met. We propose a 3-stage curriculum to train Deep Reinforcement Learning policies for bipedal walking over various challenging terrains. In the first stage, while applying forces from a target policy to the robot joints and base, the agent starts on an easy terrain and the terrain difficulty is gradually increased. In the second stage, the guiding forces are gradually reduced to zero. Finally, in the third stage, random perturbations with increasing magnitude are applied to the robot base, so the robustness of the policies are improved. In simulation experiments, we show that our approach is effective in learning separate walking policies for five terrain types: flat, hurdles, gaps, stairs, and steps. Moreover, we demonstrate that in the absence of human demonstrations, a simple hand designed walking trajectory is a sufficient prior to learn to traverse complex terrain types. In ablation studies, we show that taking out any one of the three stages of the curriculum degrades the learning performance.
\end{abstract}

%%%%%%%%%%%%%%%%%%%%%%%%%%%%%%%%%%%%%%%%%%%%%%%%%%%%%%%%%%%%%%%%%%%%%%%%%%%%%%%%
\section{Introduction}
\label{sec:introduction}

Learning to walk is difficult. Progressing from the stable locomotion of crawling to the more difficult task of walking upright takes on average 4 months for a baby. Once stable walking is achieved, the average infant falls 17 times per hour~\cite{adolph_how_2012}. For a human learning to walk, a curriculum typically occurs from crawling, to standing, to walking, and often with the guidance of an adult or walker. Many examples of curriculum learning are present throughout our lives, from progressing through school, playing sport, or learning a musical instrument~\cite{narvekar_curriculum_2020}. 

\begin{figure*}[tb!]
% \vspace{2mm}
\centering
\includegraphics[width=1.2\columnwidth]{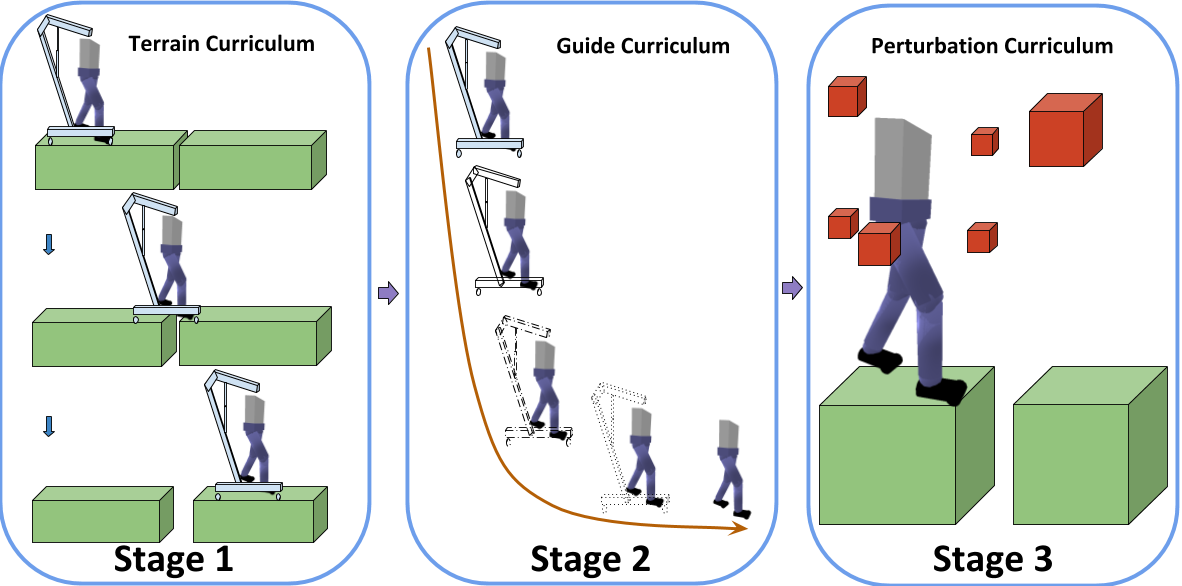}
\caption{We use curriculum learning to train a biped to traverse complex terrain samples. Our learning procedure is divided into three stages. In Stage 1 we use guiding forces that stabilise the robot Centre of Mass and encourage exploration while gradually increasing the difficulty of the terrain. In Stage 2 the guidance forces are slowly reduced. In Stage 3 we increase the magnitude of random perturbations applied to the robot to improve robustness.}
\label{fig:2}
\vspace{-3mm}
\end{figure*}

Controlling a bipedal robot over complex terrain is a challenging task. Classical control methods require meticulous hand design and tuning for each behaviour. Deep Reinforcement Learning (DRL) methods are an alternative where policies are trained through interaction with the environment, and have shown promising results for the task of walking~\cite{peng_deeploco_2017}. However, DRL relies on stochastic exploration with a single scalar reward as a learning signal, which is often difficult to design for a high degree of freedom platform. These restrictions make it difficult for policies to randomly reach the required configuration to traverse particularly challenging terrain types. Curriculum learning offers a way to alleviate these issues for difficult problem domains.

% DRL has been shown to produce policies that closely match motion capture for high degree of freedom humanoid agents, removing much of the complexity of reward design~\cite{peng_deepmimic_2018},~\cite{merel_hierarchical_2019}. These methods require motion capture examples of each behaviour. We show that we do not need perfect demonstrations to train a walking controller, and that a simple walking trajectory is a sufficient prior for a range of complex terrains.

Curriculum learning (CL) applied in machine learning has shown improved learning outcomes in shape recognition and natural language tasks~\cite{elman_learning_1993},~\cite{bengio_curriculum_2009}, with the fundamental idea that easier aspects of a task are learned first, with a gradual increase in difficulty as training progresses. CL has also been used in conjunction with DRL, with similar improvements in robotics manipulation tasks~\cite{sanger_neural_1994},~\cite{mendoza_curriculum_2017}, decision making and navigation~\cite{narvekar_curriculum_2020}, and bipedal walking ~\cite{van_de_panne_guided_1995},~\cite{wu_terrain-adaptive_2010},~\cite{yu_learning_2018}. For bipedal walking, methods that employ curriculum learning show notable reductions in training time and improved final walking performance~\cite{yu_learning_2018}. \cite{yu_learning_2018} employ external guiding forces that stablise the base of the robot, and slowly reduce these forces as a success criteria is reached (if the episode has not terminated after a fixed number of steps). By the end of training the resultant policy can operate on the target domain, free from any external guidance. Our work has similarities to~\cite{yu_learning_2018} where external forces are used to stablise the Centre of Mass (CoM) of the robot. Differently to their work, we apply external forces to each joint, provided by a rudimentary hand crafted walking trajectory, to train policies to traverse complex terrain samples. For each terrain type (flat, gaps, hurdles, stairs, and steps) we train a separate policy following the 3-stage curriculum outlined in Fig~\ref{fig:2}. 
\begin{itemize}
    \item Stage 1: Using guiding forces applied to the CoM and each robot joint, we start from an easy terrain and gradually increment the terrain difficulty
    \item Stage 2: When the terrain is at its most difficult, we slowly decrease the external guiding forces applied to the robot
    \item Stage 3: We increase the magnitude of external perturbations to improve the robustness of the policy
\end{itemize}

Our contributions are as follows:
\begin{itemize}
    \item We propose a 3-stage curriculum learning approach tailored for bipedal walking on complex terrain types
    \item We show that a single rudimentary walking trajectory can be used as a target policy to improve exploration and training outcomes for several difficult terrain types
    \item We perform extensive analysis and ablation studies showing the importance of the design choices for our curriculum approach
    \item We demonstrate the successful traversal of multiple challenging terrain types
\end{itemize}

The organization of this paper is as follows. After reviewing the relevant literature in Sec~\ref{sec:related_work}, we define the problem of interest in Sec~\ref{sec:problem} and DRL algorithm used in Sec~\ref{sec:drl}. Our curriculum learning method is presented in Sec~\ref{sec:curriculum}, and our experimental results are shown in Sec~\ref{sec:experiments}. Finally, we conclude with a brief discussion in Sec~\ref{sec:conclusion}.

%%%%%%%%%%%%%%%%%%%%%%%%%%%%%%%%%%%%%%%%%%%%%%%%%%%%%%%%%%%%%%%%%%%%%%%%%%%%%%%%
\section{Related Work}
\label{sec:related_work}

In this section, we review the literature on bipedal walking methods followed by a review of curriculum learning, predominantly in the context of deep reinforcement learning.

\subsection{Bipedal Walking}

Designing controllers for dynamic bipedal walking robots is difficult, particularly when operating over complex terrain~\cite{atkeson_what_2016}. Classical control methods for walking using Zero Moment Point (ZMP)~\cite{kajita_biped_2003} or Capture Point (CP)~\cite{pratt_capture_2006} control, rely on simplified dynamical models such as the spring loaded inverted pendulum (SLIP)~\cite{geyer_compliant_2006}. While classical methods can achieve complex behaviours~\cite{ching-long_shih_ascending_1999},~\cite{gong_feedback_2019},~\cite{xiong_bipedal_2018}, they typically require expert tuning, limiting their extension to more challenging conditions.

Deep reinforcement learning (DRL) has recently demonstrated impressive feats for bipeds in simulation~\cite{heess_emergence_2017},~\cite{peng_deeploco_2017}. Merel et al~\cite{merel_hierarchical_2019} train a humanoid to navigate and walk over gaps using an RGB camera. Policies are trained from snippets of motion capture trajectories, stitched together with a high level selection policy. Peng et al~\cite{peng_deepmimic_2018} learn complex maneuvers for high degree of freedom characters, also following motion capture data. In their work, the agent is first trained on flat terrain, then the policy is augmented to include a heightmap to allow the agent to walk up stairs. While learning from motion capture is promising, it requires having human demonstrations for all terrain conditions. We show it is possible to learn complex behaviours using a simple walking target trajectory. DRL approaches typically require extensive interaction with the target domain, which is a limiting factor for training end-to-end policies that work over a number of different terrains. In our method we train separate policies on small samples of each terrain type, such that we can add to our suite of controllers without retraining any previously trained policies.

\subsection{Curriculum Learning}

Curriculum learning (CL) has been successfully applied to a wide range of machine learning domains. The idea of CL is to provide training data to the learning algorithm in order of increasing difficulty~\cite{yu_learning_2018}. Elman~\cite{elman_learning_1993} stated the importance of starting small: learn simpler aspects of a task, then slowly increase the complexity. These ideas were applied to learning grammatical structure, by starting with a simple subset of data and gradually introducing more difficult samples~\cite{elman_learning_1993}. Bengio et al~\cite{bengio_curriculum_2009} highlight that for complex loss functions, CL can guide training towards better regions, helping find more suitable local minima. This results in a faster training time and better generalisation, demonstrated on a shape recognition task, and a natural language task. A summary of curriculum learning methods for deep reinforcement learning (DRL) can be found in Narvekar et al~\cite{narvekar_curriculum_2020}, and the blog post by Weng~\cite{weng_curriculum_2020}.

CL has been applied in continuous control domains such as robots. Sanger~\cite{sanger_neural_1994} applied Trajectory Extension Learning, where the desired trajectory for a two joint robot arm slowly moves the robot from a region of solved dynamics to a region where the dynamics are unsolved. The work by Mendoza~\cite{mendoza_curriculum_2017} shows that progressively increasing the number of controllable joints, incrementally moving robot further from the target, and reducing joint velocities improves learning on a Jaco robot arm. 

Karpathy and van de Panne~\cite{karpathy_curriculum_2012} train a dynamic articulated figure called an ``Acrobot" with a curriculum of increasingly difficult maneuvers. In their work, a low level curriculum learns a specific skill, and a high level curriculum specifies which combination of skills are required for a given task. For agents that have an unconstrained base, such as bipeds, several methods employ base stabilisation forces~\cite{van_de_panne_controller_1992},~\cite{wu_terrain-adaptive_2010},~\cite{yu_learning_2018}. Yu et al~\cite{yu_learning_2018} shows that employing a virtual assistant that applies external forces to stablise the robot base, as well as encouraging forward motion, can reduce the training time and increase the asymptotic return. The force applied by the virtual assistant is reduced as a success criteria is reached (i.e. if the robot has not fallen after a nominal number of seconds), resulting in walking and running gaits for several simulated bipedal actors~\cite{yu_learning_2018}. This work also demonstrates that with CL, a higher energy penalty can be applied to the magnitude of joint torques, without a detriment to learning. These examples are limited to simple terrain types, whereas our work investigates several complex environments.

Using a prior controller to guide exploration, and decreasing the dependence on the prior as training progresses has shown improved performance for robot navigation~\cite{rana_multiplicative_2020}. Our method combines ideas from Yu et al~\cite{yu_learning_2018}, and Rana et al~\cite{rana_multiplicative_2020}, where we stabilise the robot CoM by applying external forces, and also guide exploration using a simple prior controller applying forces to each joint. We apply these ideas, and increase the difficulty of the terrain, resulting in the traversal of complex terrain artifacts.

%%%%%%%%%%%%%%%%%%%%%%%%%%%%%%%%%%%%%%%%%%%%%%%%%%%%%%%%%%%%%%%%%%%%%%%%%%%%%%%%
% \section{Method}
% \label{sec:method}
%%%%%%%%%%%%%%%%%%%%%%%%%%%%%%%%%%%%%%%%%%%%%%%%%%%%%%%%%%%%%%%%%%%%%%%%%%%%%%%%
\section{Problem Description}
\label{sec:problem}
\begin{figure}[tb!]
% \vspace{2mm}
\centering
\includegraphics[width=0.75\columnwidth]{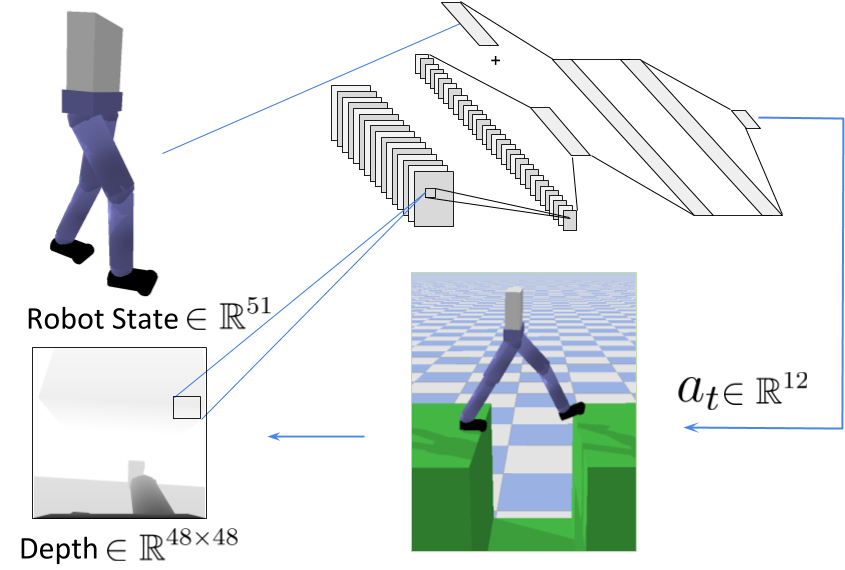}
\caption{Our policy is a neural network that takes the state (robot state and perception input) $s_t=[rs_t, I_t]$ and returns the torques applied to each joint $a_t$.}
\label{fig:network}
\vspace{-3mm}
\end{figure}

%%%%%%%%%%%%%%%%%%%%%%%%%%%%%%%%%%%%%%%%%%%%%%%%%%%%%%%%%%%%%%%%%%%%%%%%%%%%%%%%
Our biped has 12 torque controlled actuators, which we simulate in the 3D Pybullet environment~\cite{coumans_pybullet_2020}. The terrain types we investigate are flat surfaces, gaps, hurdles, stairs, and steps (shown in Fig~\ref{fig:terrains}). The state provided to each policy is $s_t=[rs_t, I_t]$, where $rs_{t}$ is the robot state and $I_t$ is the perception input at time $t$. Fig~\ref{fig:network} shows the robot state and perception input to our policy and the resulting torques that are applied to the agent in Pybullet. 

\noindent\textbf{Robot state}: $rs_t=[J_t, Jv_t, c_t, c_{t-1}, v_{CoM,t}, \omega_{CoM,t}, \\ \theta_{CoM,t}, \phi_{CoM,t}, h_{CoM,t}, s_{right}, s_{left} ]$, where $J_t$ are the joint positions in radians, $Jv_t$ are the joint velocities in rad/s, $c_t$ and $c_{t-1}$ are the current and previous contact information of each foot, respectively (four Boolean contact points per foot), $v_{CoM,t}$ and $\omega_{CoM,t}$ are the linear and angular velocities of the body Centre of Mass (CoM), $\theta_{CoM,t}$ and $\phi_{CoM,t}$ are the pitch and roll angles of the CoM and $h_{CoM,t}$ is the height of the robot from the walking surface. $s_{right}$ and $s_{left}$ are Boolean indicators of which foot is in the swing phase, and is updated when the current swing foot makes contact with the ground. All angles except joint angles are represented in the world coordinate frame. In total there are $51$ elements to the robot state, which is normalised by subtracting the mean and dividing by the standard deviation for each variable (statistics are collected as an aggregate during training).

\noindent\textbf{Perception}: Perception is provided from a depth sensor mounted to the robot base, with a resolution of $[48,48,1]$. The depth is a continuous value scaled between $0-1$, equating to a distance of 0.25 and $\SI{2}{\metre}$. The sensor moves with the body in translation and yaw, we assume we are provided with an artificial gimbal that keeps the sensor at a fixed roll and pitch. The sensor is pointed at the robot's feet (the feet are visible) and covers a distance at least two steps in front of the robot as suggested by Zaytsev~\cite{zaytsev_two_2015}. As sampling the sensor is computationally expensive, we reduce the sampling rate to $\SI{20}{\hertz}$, where the rest of the system operates at $\SI{120}{\hertz}$.

%%%%%%%%%%%%%%%%%%%%%%%%%%%%%%%%%%%%%%%%%%%%%%%%%%%%%%%%%%%%%%%%%%%%%%%%%%%%%%%%
\section{Deep Reinforcement Learning}
\label{sec:drl}
Curriculum learning in the context of reinforcement learning considers a set of tasks $i\in \mathcal{T}$~\cite{narvekar_curriculum_2020}. For each task $i$ there is a Markov Decision Process $\text{MDP}_i$, defined by tuple $\{\mathcal{S},\mathcal{A}, R, \mathcal{P}, \gamma\}$ where $s_t \in \mathcal{S}$, $a_t \in \mathcal{A}$, $r_t \in R$ are state, action and reward observed at time $t$, $\mathcal{P}$ is an unknown transition probability from $s_t$ to $s_{t+1}$ taking action $a_t$, and applying discount factor $\gamma$. We consider a task to be any instance of the curriculum, each policy will experience several tasks. For example, a task may be characterised by a particular gain $K_p$ that determines the magnitude of force applied to the robot, and the response of the environment to a particular torque will depend on the task. The goal of reinforcement learning is to maximise the sum of future rewards $R = \sum_{t=0}^{T}\gamma^tr_t$, where $r_t$ is provided by the environment at time $t$. 

% defined by tuple $\{\mathcal{S}_i,\mathcal{A}_i, R_i, \mathcal{P}_i, \gamma\}$

\subsection{Proximal Policy Optimisation (PPO)}
We choose the on-policy algorithm Proximal Policy Optimisation (PPO)~\cite{schulman_proximal_2017} to update our policy weights. As we wish to have a single policy (one for each terrain type) that evolves through several MDP's, it is preferred to use an on-policy algorithm. Actions are sampled from a deep neural network policy $a_t\sim\pi_\theta(s_t)$, where $a_t$ is a torque applied to each joint. We use the implementation of PPO from OpenAI Baselines~\cite{dhariwal_openai_2017}.

\subsection{Simple Target Policy}
\label{sec:target_policy}
We design a simple target walking trajectory with simulator physics turned off. The target trajectory is a list of joint positions that create a visually accurate walking gait of $\approx\SI{1}{\metre/\second}$. The target trajectory is divided into two segments, one for each foot, with the first set of joint positions corresponding to the impact of the respective foot. Each time the robot foot makes contact with the walking surface, we initialise the target trajectory from the appropriate segment (e.g. if the right foot contacts the ground, we restart the target trajectory with the right segment). We increment the target trajectory with each timestep.

% The target policy in the reward, and in a PD controller in the guidance curriculum (both introduced below). At each timestep we increment the position in the target walking trajectory. To improve the accuracy of our reward predictions and guidance curriculum, we link the target policy to the robot at the point where the swing foot contacts the ground. In other words, our target policy that is an open-loop set of joint positions is now linked to the robot, essentially closing the loop each time the robot swing feet change, i.e. the swing foot for the target trajectory is updated when the swing foot for the robot is updated. We replace the need for a phase variable~\cite{peng_deepmimic_2018},~\cite{merel_hierarchical_2019}, and instead use a Boolean variable that indicates which foot is in swing, and which is in stance. Each terrain type uses the same target trajectory (flat, gaps, hurdles, steps, stairs).

% The target motion could instead be a set of joint positions from motion capture, provided the motion is broken into two segments (for each swing foot), and switching between segments occurs at the same time as the robot. 

\subsection{Reward}
Our reward is inspired by Peng et al in their work mimicking motion capture data~\cite{peng_deepmimic_2018}. Our reward function is the same for each policy, defined as:
\begin{equation}
r_t = goal + pos + vel + base + step + act
\end{equation}
where:
\begin{itemize}[leftmargin=*]
    \item $goal$ is the error between the current heading velocity and goal heading velocity $= w_{goal}\cdot \exp[-c_{goal}\cdot e_{goal}]$.
    \item $pos$ is the error between the joint positions and the target joint positions $= w_{pos}\cdot \exp[-c_{pos}\cdot e_{pos}]$.
    \item $vel$ is the error between the current joint velocities and the target joint velocities (target joint velocities are set to zero) $= w_{vel}\cdot \exp[-c_{vel}\cdot e_{vel}]$.
    \item $base$ is the CoM error from a target height, and zero roll and pitch positions $= w_{base}\cdot \exp[-c_{base}\cdot e_{base}]$. 
    \item $step$ encourages a symmetrical step length ($\text{left}_{step}\approx \text{right}_{step}$ when in contact) $= w_{step}\cdot \exp[-c_{step}\cdot e_{step}]$. 
    \item $act$ is a penalty for excessive torque for each of the $n$ joints $= w_{act}\sum_n(a_t^2)$. 
\end{itemize}

$e$ is the error of each respective element. Parameters $w$ and $c$ are tunable weights, which are fixed for all policies. 

%%%%%%%%%%%%%%%%%%%%%%%%%%%%%%%%%%%%%%%%%%%%%%%%%%%%%%%%%%%%%%%%%%%%%%%%%%%%%%%%
\section{Curriculum Learning}
\label{sec:curriculum}
We design our curriculum with the rationale that dynamic platforms are usually supported (e.g. with a crane and harness) while controllers are developed. The learning process is divided into three stages, where $\mathcal{D}$ is the terrain curriculum (Stage 1), $\mathcal{F}$ is the guide curriculum (Stage 2), and $\mathcal{P}$ is the perturbation curriculum (Stage 3), where each stage represents a set of difficulties $\mathcal{D} = \{d_1, d_2, .., d_n\}$, $\mathcal{F} = \{f_1, f_2, .., f_n\}$, and $\mathcal{P} = \{p_1, p_2, .., p_n\}$. To begin learning we select $d_1, f_1$ and $p_1$, i.e all curriculum stages are at the lowest difficulty level. We then proceed to increment the difficulty of Stage 1 to its most difficult setting ($d_1 \rightarrow d_n$), then Stage 2 ($f_1 \rightarrow f_n$), and finally Stage 3 ($p_1 \rightarrow p_n$). During policy training, each increment to the next difficulty level occurs with a \textbf{success criteria}, which we define as the robot successfully reaching the end of the terrain sample three consecutive times. 

\subsection{Stage 1. Terrain Curriculum:}
In the terrain curriculum $\mathcal{D} = \{d_1, d_2, .., d_n\}$, the terrain difficulty is increased by incrementing $d$ up to a final difficulty $n$. In this stage, we apply the guide curriculum and perturbation curriculum at their lowest difficulty. The difficulty range  of each terrain is shown in Fig~\ref{fig:terrains}. Increases to the terrain difficulties are linear, incrementing terrain dimensions by a fixed amount each time the success criteria is achieved. 

% The terrain curriculum allows us to increase the difficulty of the terrain, where for some terrains (like hurdles) the policy fails to traverse the terrain at all, or for others (like gaps) the policy can take much longer to learn the required behaviour.

\subsection{Stage 2. Guide Curriculum:}
\label{sec:guide_curriculum}
Guide forces are employed from the beginning of training, tracking the target policy explained in Sec.~\ref{sec:target_policy}. Once the terrain curriculum has completed we begin the guide curriculum $\mathcal{F} = \{f_1, f_2, .., f_n\}$. Guide forces are applied to the CoM of the robot $f_c$ and to each joint $f_j$, where $f_1 = [f_c, f_j]$ is set to the lowest difficulty (highest guide forces).

$f_c$ is a PD controller that is applied to all 6 degrees of freedom of the CoM. 
\begin{equation}
f_c = K_{p_c} (p_{target} - p_{CoM}) + K_{d_c} ( v_{target} - v_{CoM})
\end{equation}
Where $K_{p_c}$ and $K_{d_c}$ are the gains, $p_{target}$ is the target position, $p_{CoM}$ is the observed position, $v_{target}$ is the target velocity, and $v_{CoM}$ is the observed velocity of the robot CoM. The target for the velocity in the forward $x$ direction is $\SI{1}{\meter/\second}$. The height $z$ target position is fixed at a nominal height, the $yaw$ targe follows the yaw of the terrain, and the position targets for the remaining degrees of freedom $y, roll, pitch$, are zero. All velocity targets, except for $x$ velocity, are set to zero. We liken this idea to fixing the robot to a small crane or gantry that supports the body of the robot, and moves forward at the desired velocity.
% This approach has shown reduced training times and increased asymptotic performance for multiple morphologies with a floating base~\cite{yu_learning_2018}.

$f_j$ is a second PD controller that is applied to all 12 actuators tracking our simple target trajectory. Forces applied to each joint are calculated from:
\begin{equation}
f_j = K_{p_j}( J_{t_{target}} - J_t ) + K_{d_j}(Jv_{t_{target}} - Jv_t )
\end{equation}
Where $K_{p_j}$ and $K_{d_j}$ are the gains, $J_{t_{target}}$ is the joint positions of the target trajectory, $J_t$ is the observed joint positions, $Jv_{t_{target}}$ is the target joint velocities, and $Jv_t$ is the observed joint velocities. We found that setting the target velocities to zero was sufficient for training (rather than estimating the target velocities with finite differences).

The guide curriculum involves reducing the guide forces in discrete steps following: $f_2 = 0.65 \cdot f_1$. The external forces are not provided as input to policy, and are considered an artifact of the environment. Once the guide forces decrease below a threshold ($f_n$) Stage 2 is complete.

\subsection{Stage 3. Perturbation Curriculum:}
% \subsection{Stage 3. Adding Perturbations}

Random perturbations applied during training have been shown to improve the robustness of walking policies~\cite{schulman_proximal_2017}. When the other curriculum stages are complete, we increase the magnitude of external perturbations with the perturbation curriculum, $\mathcal{P} = \{p_1, p_2, .., p_n\}$. Perturbations are an impulse force of random magnitude, applied to all 6 degrees of freedom of the CoM. At a random interval of approximately 2.5Hz, and for each DOF, we sample a perturbation force from a uniform distribution $(-p_1,p_1)$, where $p_1$ is increased linearly to the final value $p_n$.

%%%%%%%%%%%%%%%%%%%%%%%%%%%%%%%%%%%%%%%%%%%%%%%%%%%%%%%%%%%%%%%%%%%%%%%%%%%%%%%%
\section{Experiments}
\label{sec:experiments}

% \begin{figure*}[tb!]
% % \vspace{2mm}
% \centering
% \subfloat[Curved flat terrain with a bend angle starting at $20\degree$ and increasing to $120\degree$]{\includegraphics[width=\wI cm, height=\hI cm]{images/flat.png}}
% \hfill
% \subfloat[Hurdles with a height starting at $\SI{6}{\cm}$ and increasing to $\SI{30}{\cm}$]{\includegraphics[width=\wI cm, height=\hI cm]{images/jumps.png}}
% \hfill
% \subfloat[Gaps with a width starting at $\SI{7}{\cm}$ and increasing to $\SI{70}{\cm}$]{\includegraphics[width=\wI cm, height=\hII cm]{images/compare_gaps.png}}
% \hfill

% \caption{Using a single walking trajectory as a guide, our curriculum method traverses complex terrain types. The initial and final terrain sizes are provided.}
% \label{fig:terrains}
% \vspace{-3mm}
% \end{figure*}

\begin{figure}[tb!]
% \vspace{2mm}
\centering
\subfloat[Curved flat terrain initial angle starting at $20\degree$ and width $\SI{1.1}{\meter}$ ]{\includegraphics[width=\w cm, height=\h cm]{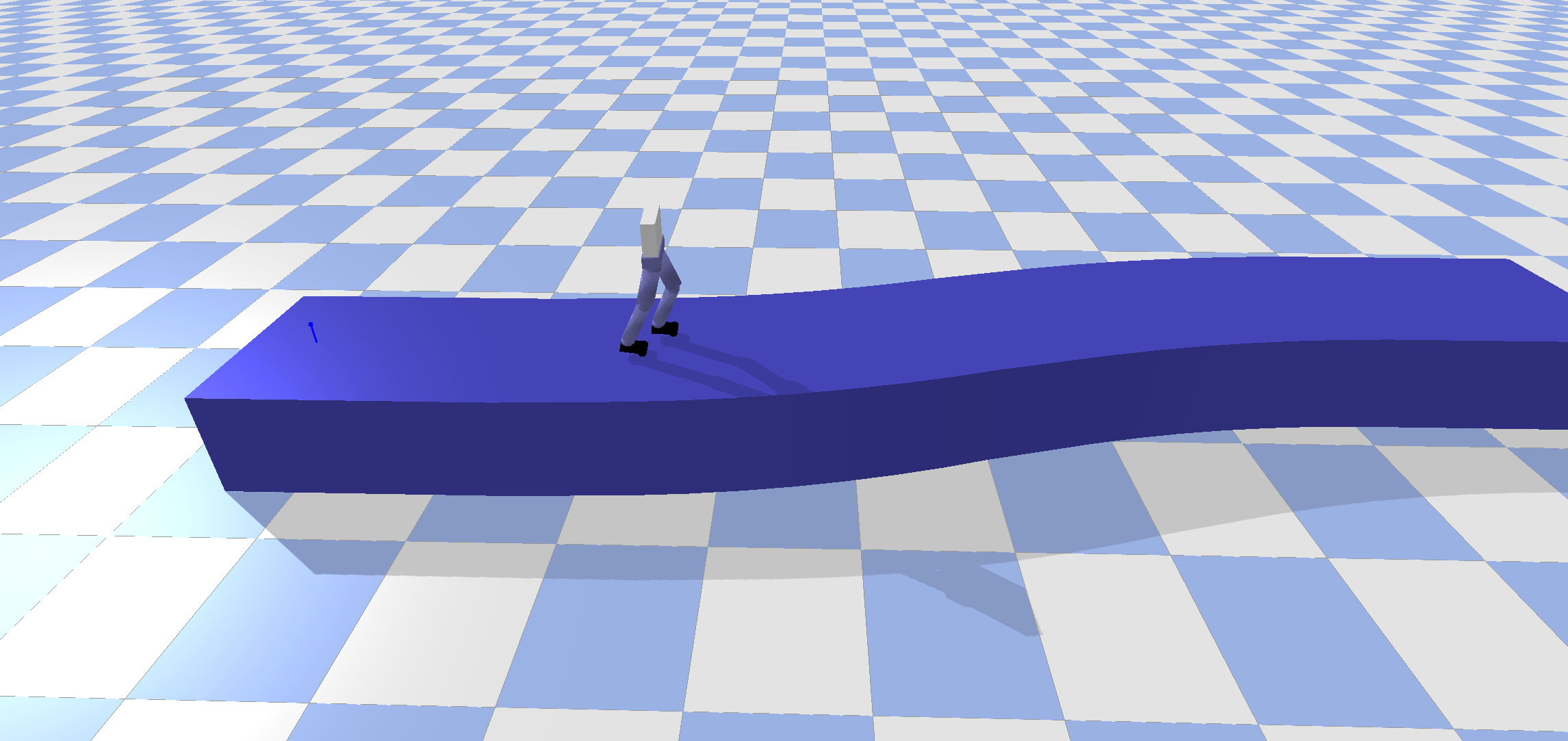}}
\hfill
\subfloat[Curved flat terrain final angle
$120\degree$ and width $\SI{0.9}{\meter}$]{\includegraphics[width=\w cm, height=\h cm]{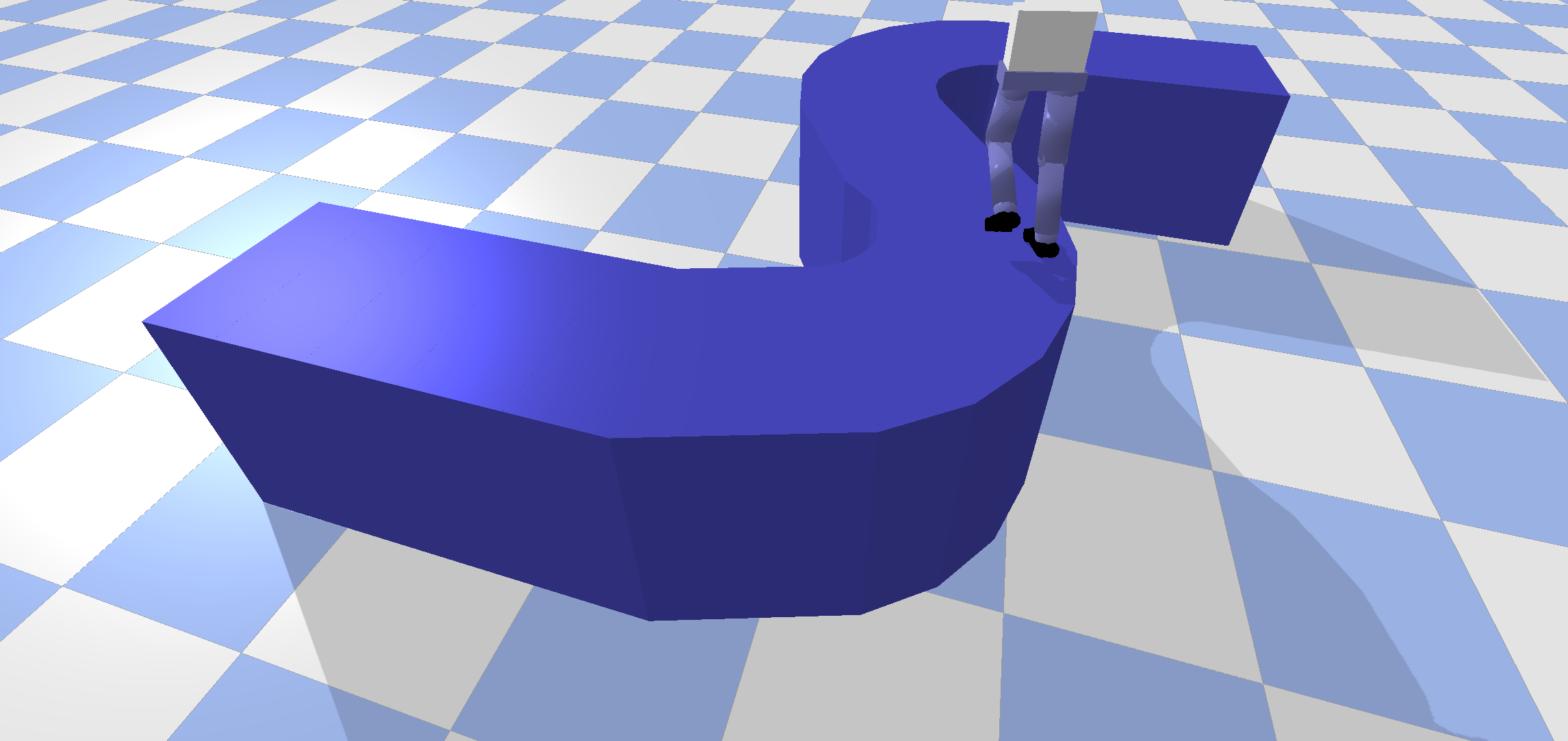}}
\hfill
\subfloat[Hurdles initial height $\SI{13}{\cm}$]{\includegraphics[width=\w cm, height=\h cm]{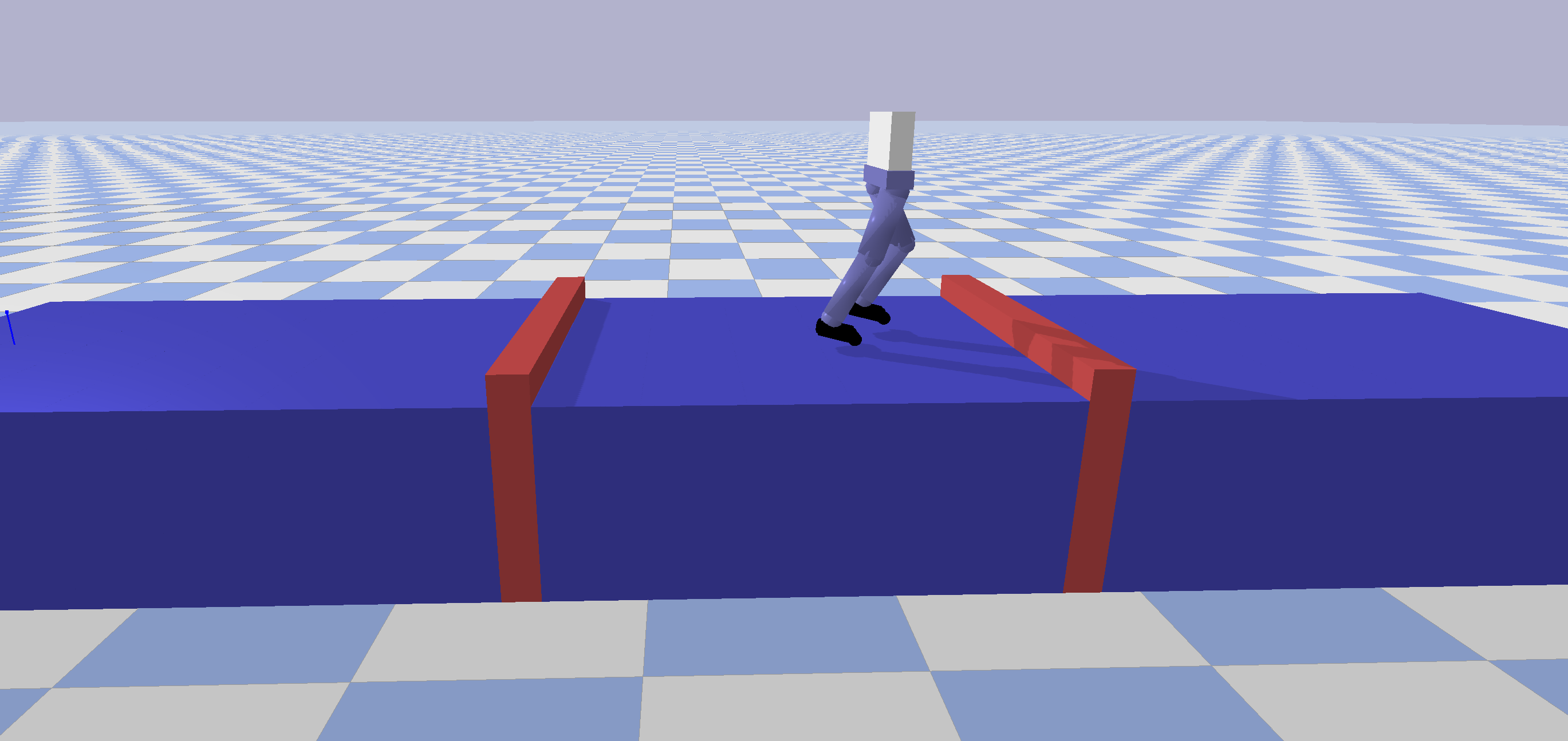}}
\hfill
\subfloat[Hurdles final height $\SI{38}{\cm}$]{\includegraphics[width=\w cm, height=\h cm]{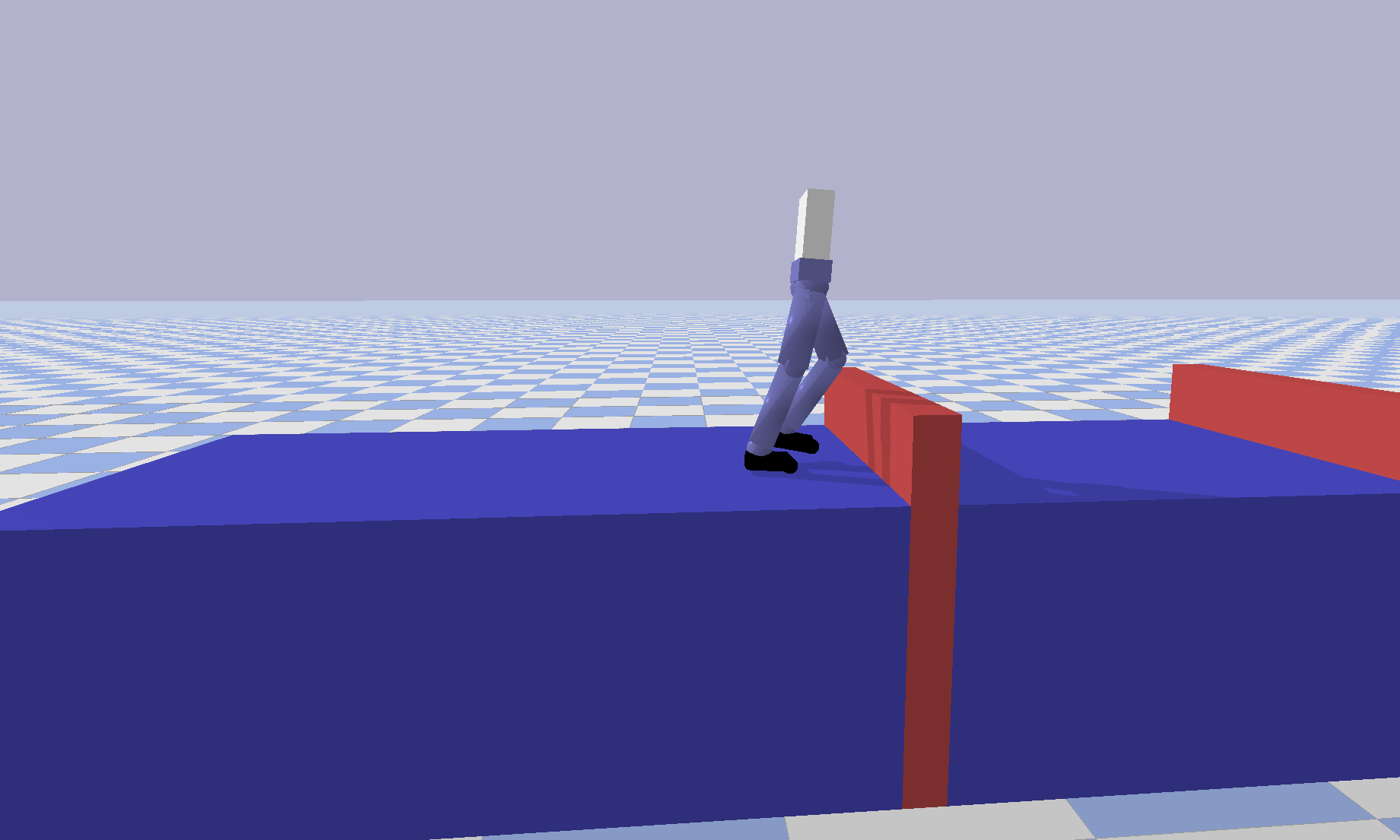}}
\hfill
\subfloat[Gaps initial length $\SI{10}{\cm}$]{\includegraphics[width=\w cm, height=\h cm]{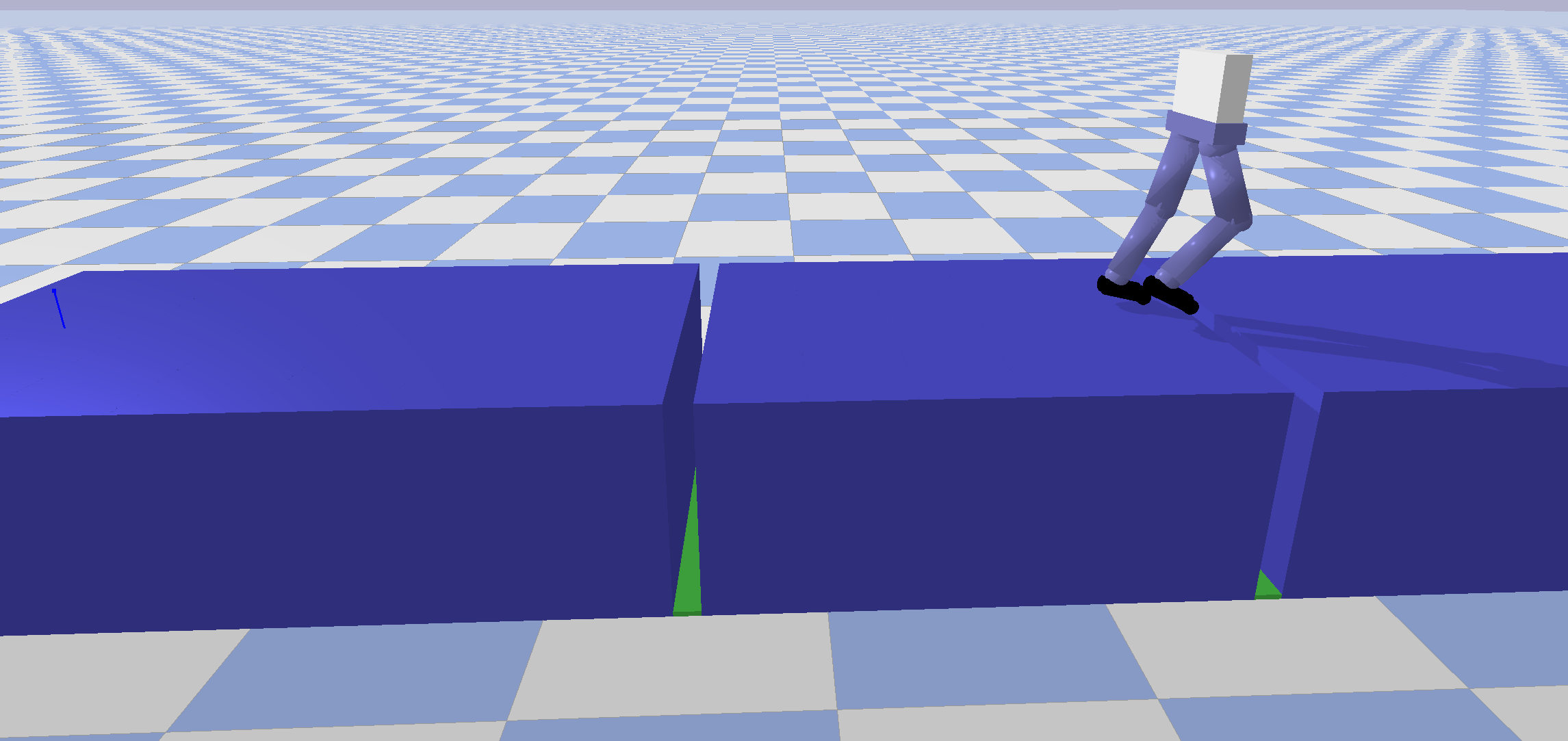}}
\hfill
\subfloat[Gaps final length $\SI{100}{\cm}$]{\includegraphics[width=\w cm, height=\h cm]{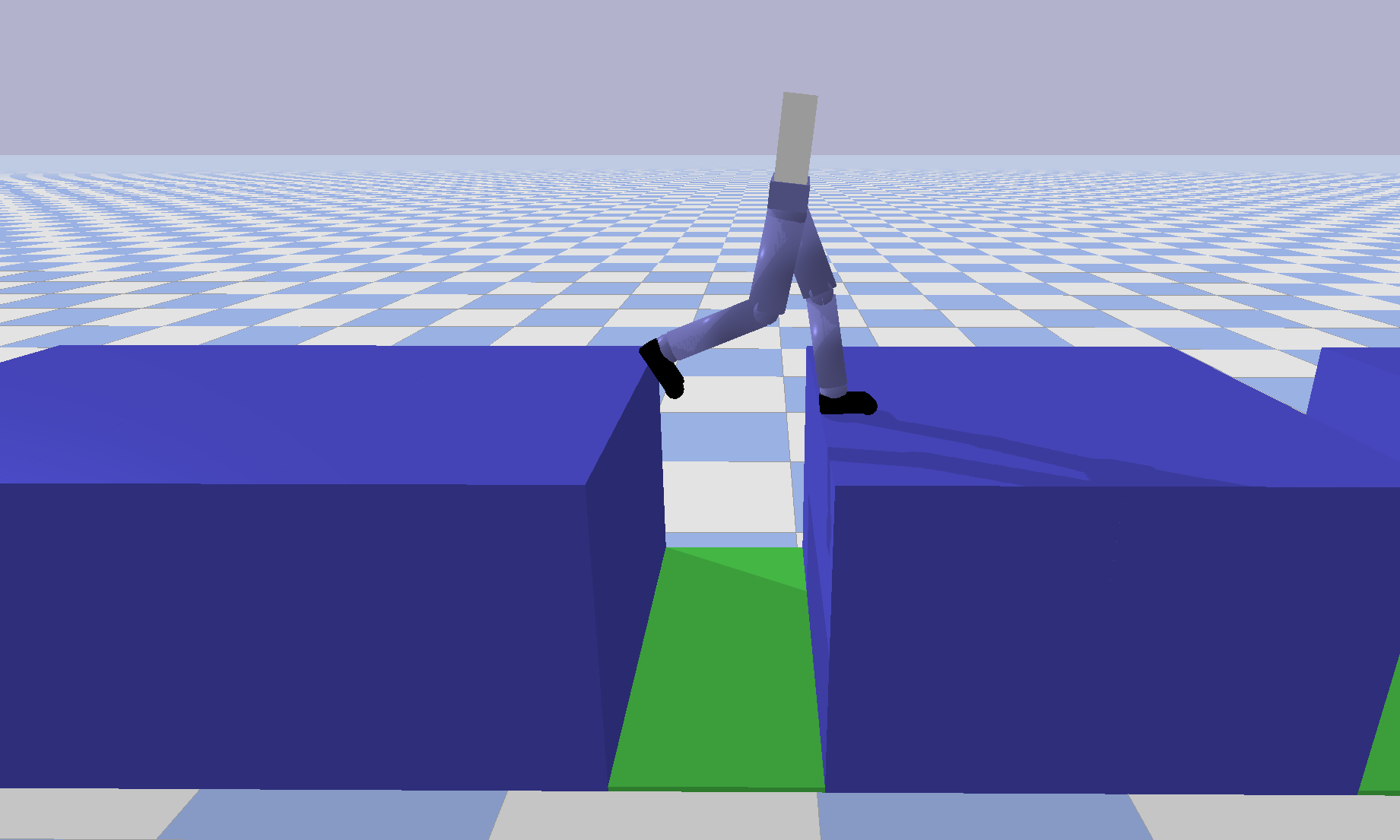}}
\hfill
\subfloat[Stairs initial height $\SI{1.7}{\cm}$]{\includegraphics[width=\w cm, height=\h cm]{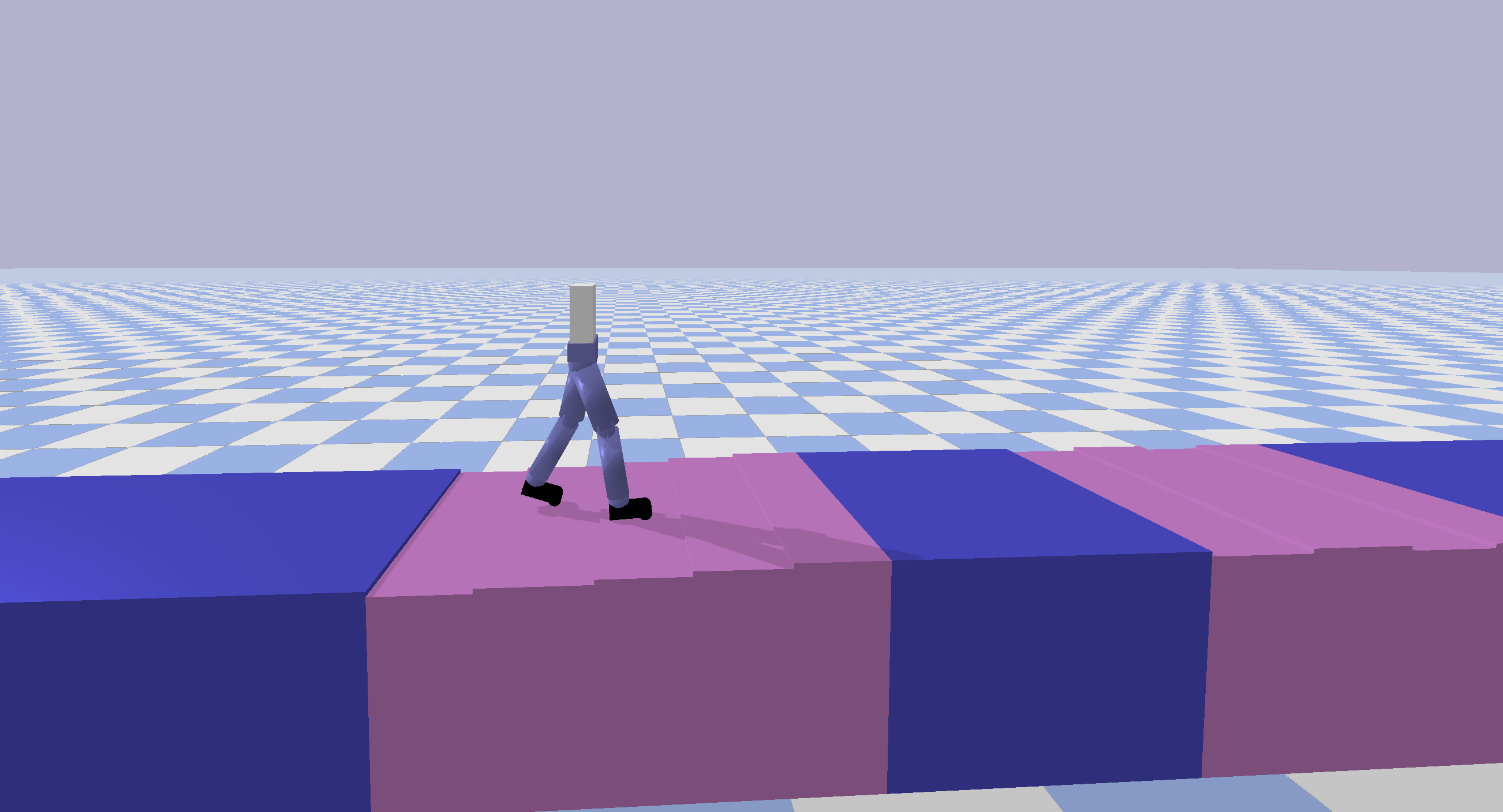}}
\hfill
\subfloat[Stairs final height $\SI{17}{\cm}$]{\includegraphics[width=\w cm, height=\h cm]{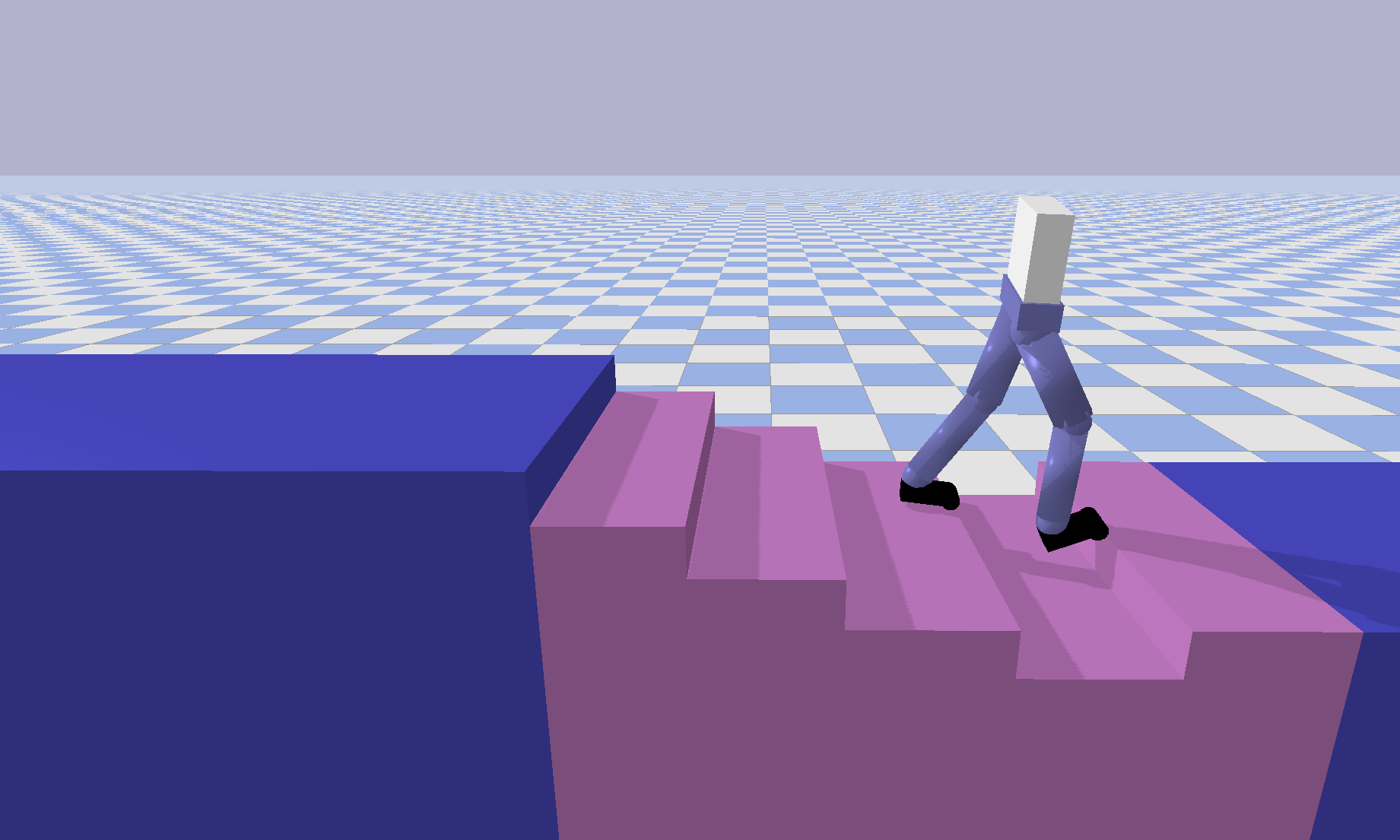}}
\hfill
\subfloat[Steps initial distance apart $\SI{4}{\cm}$]{\includegraphics[width=\w cm, height=\h cm]{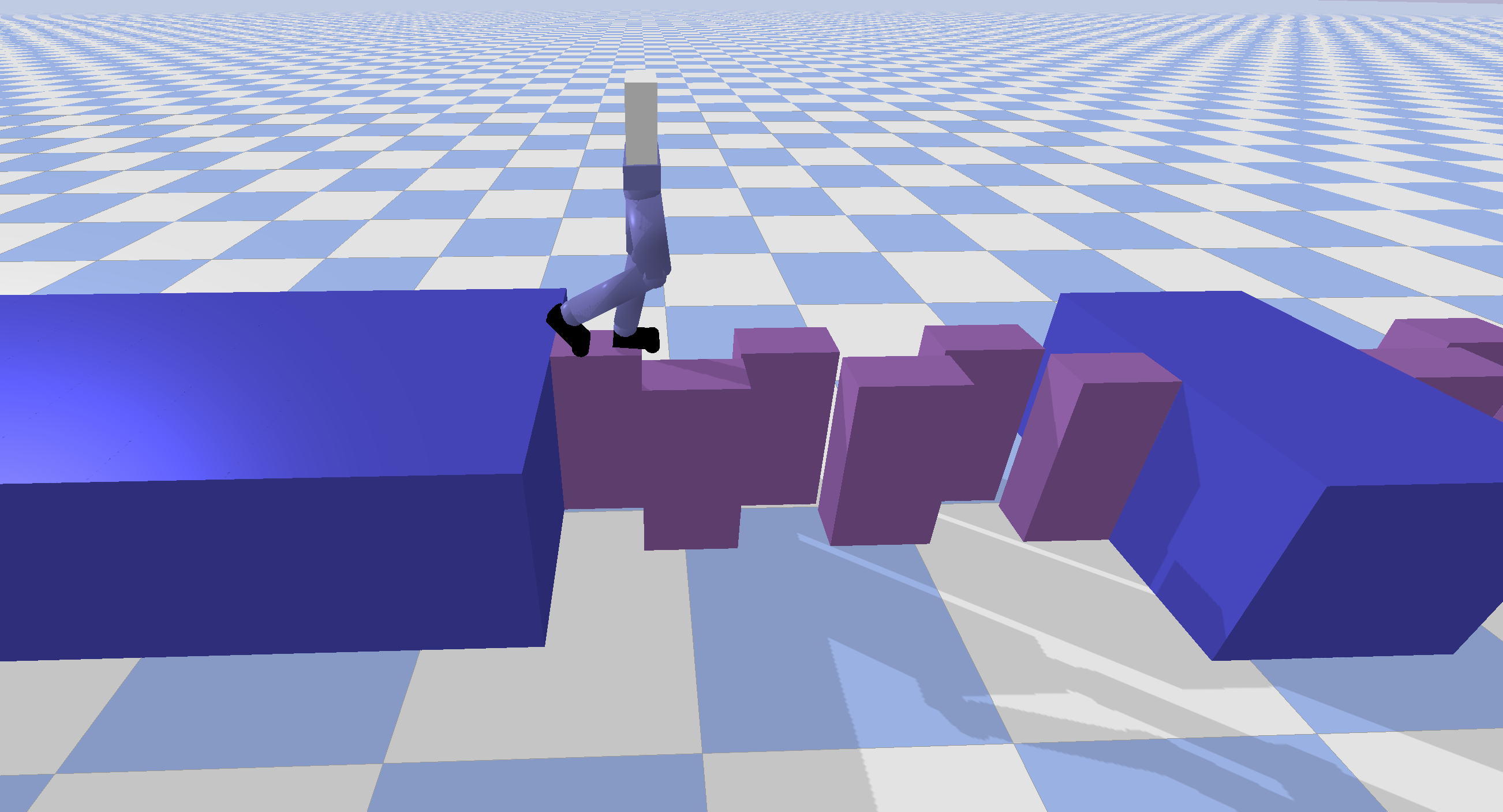}}
\hfill
\subfloat[Steps final distance apart $\SI{40}{\cm}$]{\includegraphics[width=\w cm, height=\h cm]{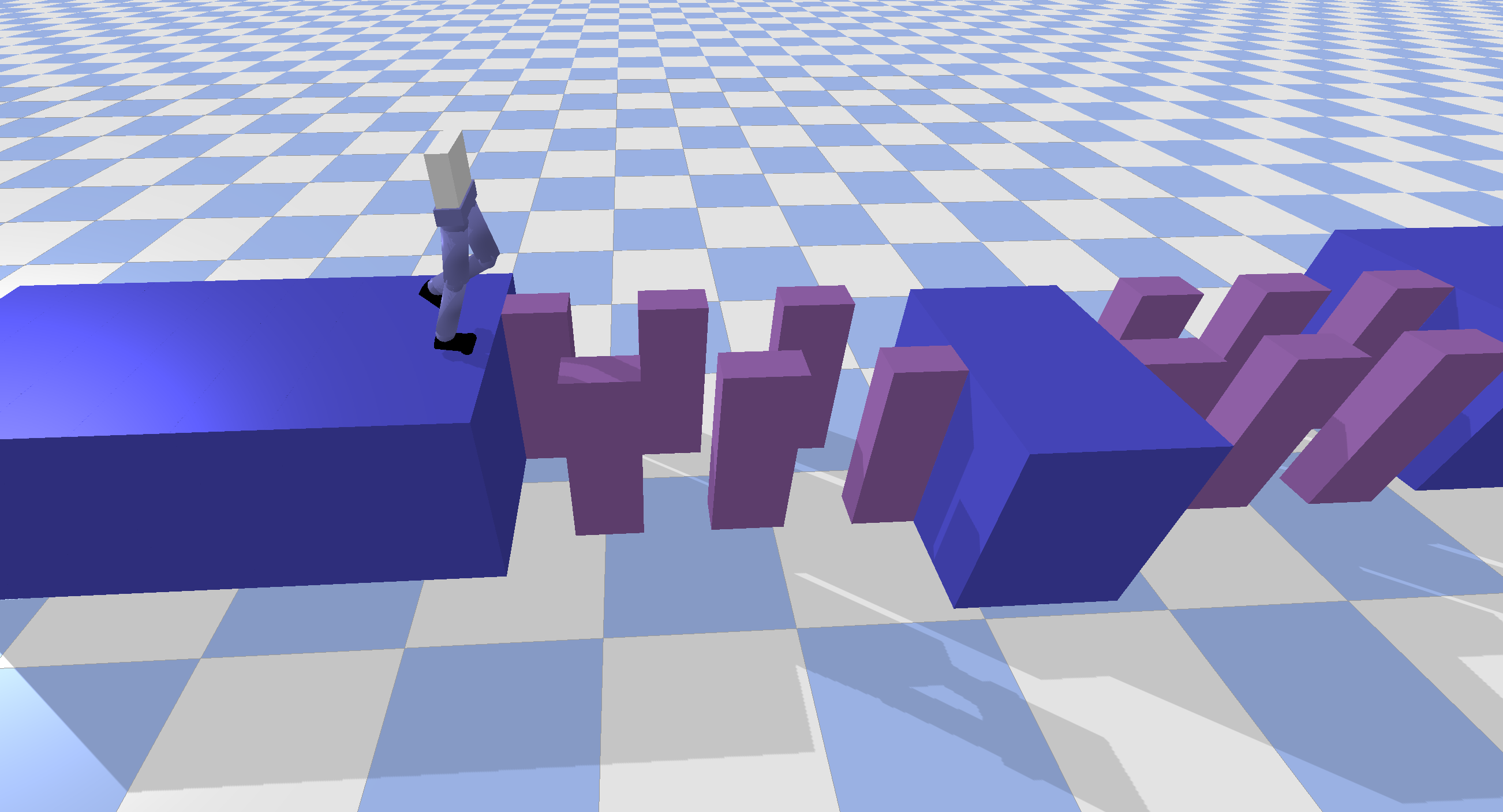}}
% \hfill
\caption{Our curriculum method traverses complex terrain types. The initial and final terrain sizes used in training are provided.}
\label{fig:terrains}
\vspace{-3mm}
\end{figure}

In this section we introduce the terrain types used for the test worlds (Sec~\ref{sec:terrain_types}), and present our simulation results. We evaluate Stage 1 (terrain curriculum), and Stage 2 (guide curriculum) of our method in Sec~\ref{sec:stage1_stage2}, before analysing Stage 3 (perturbation curriculum) in Sec~\ref{sec:stage3}. 

\subsection{Terrain Types}
\label{sec:terrain_types}
For all terrains there is a total of 10 steps of evenly spaced difficulty ($d_1 \rightarrow d_{10}$). The suite of terrains and range of difficulties is shown in Fig~\ref{fig:terrains}. Terrain conditions are selected to test our system with a broad range of scenarios. Examples include the Flat terrain that has curves that require the robot to turn (Fig~\ref{fig:terrains}(a-b)), the Hurdles terrain (Fig~\ref{fig:terrains}(c-d)) that forces the robot to take a high step, and the Steps terrain (Fig~\ref{fig:terrains}(i-j)) where the robot must step with a lateral component.
% Movements such as turning, high steps, or lateral stepping are not present in the simple target trajectory, yet our policies adapt to the requirements of the terrain. 

% Each terrain example requires the robot to use its perception, this is enforced by randomisation of the starting location of the robot, and randomised aspects of terrain generation (direction of the curve, number of steps or stairs). 

Each policy is trained on a scenario with two consecutive instances of the same terrain artifact, where an instance refers to a single gap or hurdle, or a random number ($5-8$) of stairs or steps. We found that training on two consecutive artifacts performed better during evaluation than a sample with a single artifact. We evaluate all methods on a sequence of seven consecutive instances of the same terrain artifact at maximum difficulty, making up one trial. We run each policy for 500 trials and record the distance travelled as a percentage of total distance, which we use as our evaluation metric. For all experiments no external disturbance forces are applied when evaluating a policy, except for the perturbation evaluation section of Sec~\ref{sec:intense}, where perturbations are applied during evaluation. For all experiments, the terrain difficulty at evaluation is set to the highest difficulty (10), except for the domain difficulty section of Sec~\ref{sec:diff}, where we evaluate our policies on a range of difficulties.

\subsection{Stage 1 and Stage 2 Analysis}
\label{sec:stage1_stage2}

\begin{figure*}[ht]
\vspace{-2mm}
\centering
\subfloat[Flat]{\includegraphics[width=.3\textwidth, height=\hIII cm]{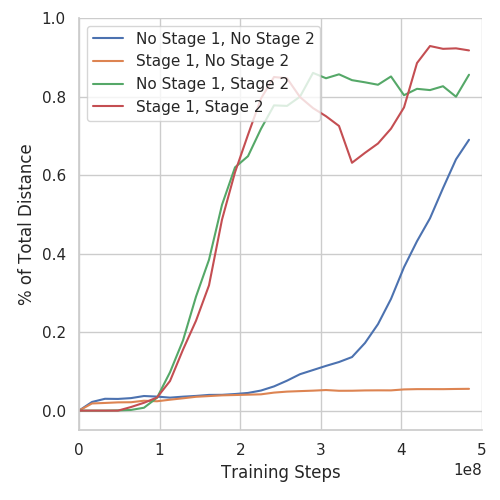}}\quad
\subfloat[Gaps]{\includegraphics[width=.3\textwidth, height=\hIII cm]{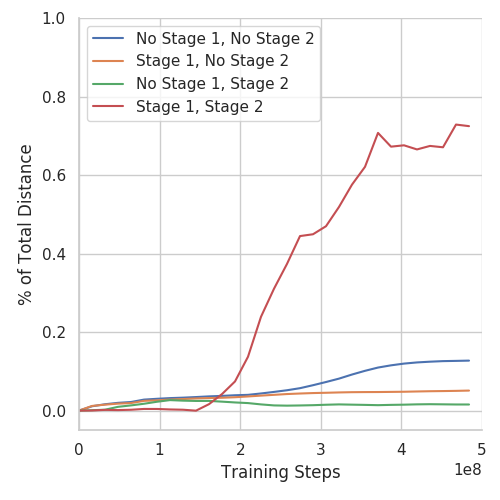}}\quad
\subfloat[Hurdles]{\includegraphics[width=.3\textwidth, height=\hIII cm]{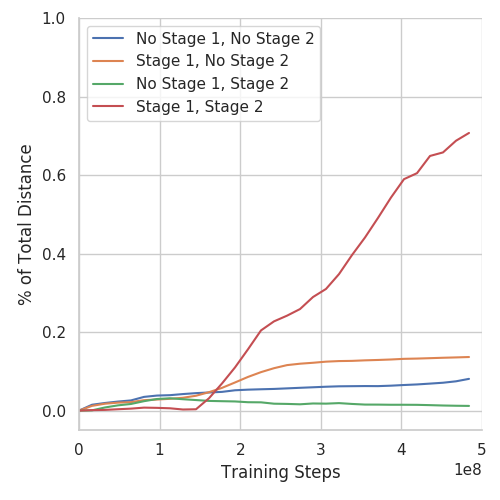}}

\medskip
\vspace{-5mm}
\subfloat[Stairs]{\includegraphics[width=.3\textwidth, height=\hIII cm]{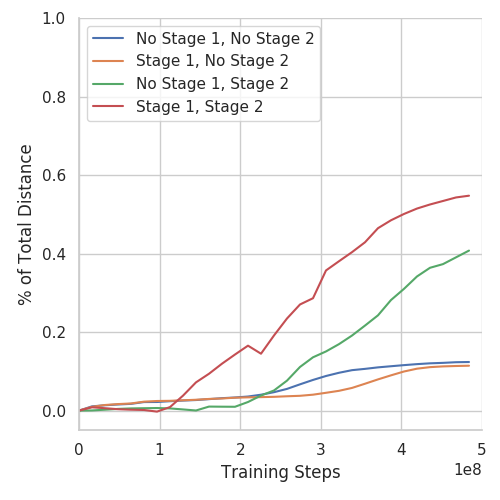}}\quad
\subfloat[Steps]{\includegraphics[width=.3\textwidth, height=\hIII cm]{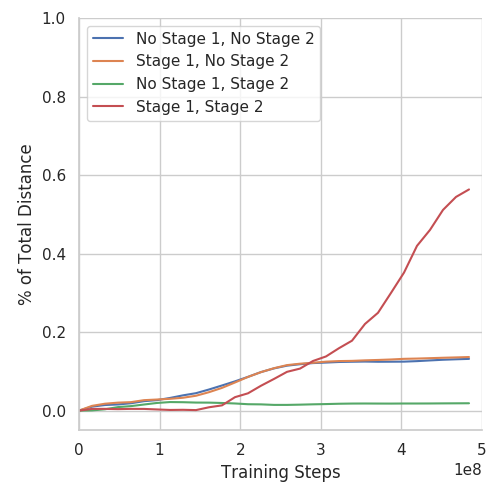}}

\caption{Percentage of total distance travelled, evaluated during training showing the comparisons of the Stage 1 and Stage 2 ablation study.}
\label{fig:compare}
\end{figure*}
\begin{table}[h!]
    \vspace{-2mm}
    % \vspace{-45.5mm}https://www.overleaf.com/project/5efc80ce9e222200014e6c84
    \centering
    \begin{adjustbox}{max width=1.0\columnwidth}  
    \begin{tabular}{cccccc}
        %  \multicolumn{1}{c|}{\textbf{Stage 1 and Stage 2 Analysis}} & Flat & Gap & Hurdle & Stairs & Steps \\
          & Flat & Gap & Hurdle & Stairs & Steps \\
         \hline
         No Stage 1, No Stage 2       & 79.4           & 12.8          & 10.3          & 12.4          & 13.4 \\
         Stage 1, No Stage 2          & 5.5  & 5.3          & 7.1          &  11.5          & 14.2 \\
         No Stage 1, Stage 2          & 88.7           &  1.5          &  1.2          &  43.9          &   1.9 \\
         Stage 1, Stage 2     &  \textbf{89.8}            & \textbf{69.5}           &  \textbf{79.0}          & \textbf{55.8} & \textbf{58.1}   \\
    \hline
    \end{tabular}
    \end{adjustbox}
    \caption{Ablation study of Stage 1 and Stage 2 showing percentage of total distance travelled, computed with 500 trials.}
    \label{tab:results_ablation}
    \vspace{-5mm}
\end{table}

This section analyses the effect of Stage 1 and Stage 2 on policy performance in an ablation study. We also investigate three individual aspects of our method: the effect of using success criteria to decay forces for Stage 2, removing the guiding forces applied to the joints, and not linking the target policy to the robot state. For all experiments in this section, the external perturbations are set at the lowest level $p_1$ throughout training, with no external perturbations applied during evaluation.

\subsubsection{Ablation study of Stage 1 and Stage 2}
We investigate the benefit of Stage 1 and Stage 2 of our method by removing one or both stages.

\begin{itemize}[leftmargin=*]

\item\textbf{No Stage 1, No Stage 2:}  The terrain starts at maximum difficulty. Guiding forces are not applied to the robot CoM or joints.

\item\textbf{Stage 1, No Stage 2:} The terrain curriculum is active. Guiding forces are not applied to the robot CoM or joints.

\item\textbf{No Stage 1, Stage 2:} The terrain starts at maximum difficulty. Guide curriculum is active.

\item\textbf{Stage 1, Stage 2:} Both the terrain curriculum and guide curriculum are active. 
\end{itemize}

Fig~\ref{fig:compare} shows the evolution of the percentage of total distance achieved with respect to the number of training steps. We perform this evaluation periodically by pausing training, removing any external forces, setting the terrain difficulty to maximum, and running 500 trials. The final evaluation is provided in Table~\ref{tab:results_ablation}. We can see from Fig~\ref{fig:compare}, and Table~\ref{tab:results_ablation} that both Stage 1 and Stage 2 are important for the development of successful policies for the selected terrain types. Stage 1 is necessary for all terrains except for Flat and Stairs. \textbf{No Stage 1, Stage 2} on the Stairs terrain results in the traversal of a significant portion of the test world (43.9\%). Without Stage 2 the agent learns a one leg hopping gait. For all terrains this gait is unstable (the agent is unable to balance beyond the first hop), except for \textbf{No Stage 1, No Stage 2} on the Flat terrain where the agent successfully traverses 79.4\% of the test world by hopping on one leg. The lowest performance is found with \textbf{No Stage 1, Stage 2} for the Gap, Hurdle, and Steps terrains. For these terrain types, without the terrain curriculum the robot is unable to reach the success criteria required to lower the guidance forces. When evaluated without guide forces, the robot is unable to balance and achieves a much lower result than without any curriculum at all.

% (Flat 89.8\%, Gaps 69.5\%, Hurdles 79.0\%, Stairs 55.8\%, Steps 58.1\%)

% , for each terrain type (evaluated on 7 consecutive instances)

% \begin{table*}[t]
%     % \vspace{-45.5mm}
%     \centering
%     \begin{tabular}{cccccc}
%         %  \multicolumn{1}{c|}{\textbf{Stage 1 and Stage 2 Analysis}} & Flat & Gap & Hurdle & Stairs & Steps \\
%           & Flat & Gap & Hurdle & Stairs & Steps \\
%          \hline
%          No Stage 1, No Stage 2       & 91.1           & 12.7          & 13.9          & 14.9          & 13.5 \\
%          Stage 1, No Stage 2          & \textbf{95.7}  & 14.0          & 14.3          &  3.8          & 15.3 \\
%          No Stage 1, Stage 2          & 89.8           &  1.9          &  2.7          &  1.1          &  1.2 \\
%          Stage 1, Stage 2     &        92.4            & \textbf{78.9} & \textbf{71.6} & \textbf{38.3} & \textbf{60.8}  \\
%          Continuous Stage 2 decay   &  79.1            & 63.9          & 14.2          & 33.5           & 56.7  \\
%          Guide forces on CoM only   &  47.4            &  2.8          &  2.4          & 25.6          &  1.0 \\
%          No link to target          &    7.7           & 54.1          & 53.7          & 17.6          &  14.6 \\
%     \hline
%     \end{tabular}
%     \caption{Analysis of Stage 1 and Stage 2 showing percentage of total distance travelled, computed with 500 trials.}
%     \label{tab:results1}
% \end{table*}

\subsubsection{Effect of using success criteria to decay forces}
\begin{table}[h!]
    % \vspace{-45.5mm}
    \vspace{-2mm}
    \centering
    \begin{adjustbox}{max width=1.0\columnwidth}  
    \begin{tabular}{cccccc}
        %  \multicolumn{1}{c|}{\textbf{Stage 1 and Stage 2 Analysis}} & Flat & Gap & Hurdle & Stairs & Steps \\
          & Flat & Gap & Hurdle & Stairs & Steps \\
         \hline
         Force decay with success   &   89.8            & \textbf{69.5}           &  \textbf{79.0}          & \textbf{55.8} & 58.1  \\  
         Continuous force decay   &  \textbf{97.7}            & 15.2          & 22.8          & 38.7           & \textbf{58.3}  \\
    \hline
    \end{tabular}
    \end{adjustbox}
    \caption{Comparison of force decay used in Stage 2 as percentage of total distance travelled, computed with 500 trials.}
    \label{tab:results_decay}
    \vspace{-2mm}
\end{table}
This section investigates the choice of decay on guidance forces (Stage 2). We compare our method of decay that is dependent on agent success, to decaying the guiding forces at fixed intervals irrespective of the agent success. We found that decaying the guiding forces by $0.995$ each episode was a similar rate to our decay approach (introduced in Sec~\ref{sec:guide_curriculum}). The results in Table ~\ref{tab:results_decay} show that for the Gaps, Hurdles, and Stairs terrain types, decaying the guide forces continuously performed significantly worse on the test world than our method (Gaps 15.2\% compared to 69.5\%, Hurdles 22.8\% compared to 79.0\%, Stairs 38.7\% compared to 55.8\%). For the Flat and Steps terrain types the continuous decay performed slightly better than decay based on agent success (Flat 89.8\% compared to 97.7\%, Steps 58.1\% compared to 58.3\%). We suspect a decay rate tuned for each terrain type would match or improve the performance of our method for all terrains, though introduces undesirable hand tuning.

\subsubsection{Effect of guide forces on CoM only}
\begin{table}[h!]
    % \vspace{-45.5mm}
    \vspace{-2mm}
    \centering
    \begin{adjustbox}{max width=1.0\columnwidth}  
    \begin{tabular}{cccccc}
        %  \multicolumn{1}{c|}{\textbf{Stage 1 and Stage 2 Analysis}} & Flat & Gap & Hurdle & Stairs & Steps \\
          & Flat & Gap & Hurdle & Stairs & Steps \\
         \hline
        Forces on CoM and joints   &   \textbf{89.8}            & \textbf{69.5}           &  \textbf{79.0}          & \textbf{55.8} & \textbf{58.1}  \\
         Forces on CoM only   &  81.6            &  8.6          &  2.7          & 12.2          & 1.0 \\
    \hline
    \end{tabular}
    \end{adjustbox}
    \caption{Effect of applying guide forces to CoM only, shown as a percentage of total distance travelled, computed with 500 trials.}
    \label{tab:results_guide}
    \vspace{-2mm}
\end{table}
We test the effect of removing the guiding forces on the joints, instead only apply forces to the CoM during Stage 2 Guide Curriculum. We found guiding forces applied only to the base performed worse on all terrains than also applying guiding forces to the joints (Table~\ref{tab:results_guide}). Notably low scores (Gaps 8.6\%, Hurdle 2.7\%, Steps 1.0\%) suggest the agent was unable to decay the guide forces on the CoM, so unable to support itself during evaluation.

\subsubsection{Effect of not linking target policy}
\begin{table}[h!]
    \vspace{-2mm}
    \centering
    \begin{adjustbox}{max width=1.0\columnwidth}    
    \begin{tabular}{cccccc}
        %  \multicolumn{1}{c|}{\textbf{Stage 1 and Stage 2 Analysis}} & Flat & Gap & Hurdle & Stairs & Steps \\
          & Flat & Gap & Hurdle & Stairs & Steps \\
         \hline
         Link to Target Policy   &   \textbf{89.8}            & \textbf{69.5}           &  \textbf{79.0}          & \textbf{55.8} & \textbf{58.1}  \\
         No link to Target Policy          &    54.6           & 38.9          & 40.9          & 28.8          &  15.2 \\
    \hline
    \end{tabular}
    \end{adjustbox}
    \caption{Effect of linking the target trajectory to the robot, shown as a percentage of total distance travelled, computed with 500 trials.}
    \label{tab:results_no_link}
    \vspace{-2mm}
\end{table}
We investigate the effect of removing the link between the target policy and the robot. Linking is introduced in Sec~\ref{sec:target_policy}, and refers to resetting the target trajectory with the respective segment (left or right) each time a foot of the robot makes contact with the ground. When the target trajectory is not linked to the robot, the target joint positions are initialised only at the first step. Advancing the target trajectory occurs with each timestep, without any feedback from the robot. We found that not linking the target trajectory with the robot reduced the performance of all terrains in comparison to our method as seen in Table~\ref{tab:results_no_link}. While some policies were able to successfully traverse portions of the test world (Flat 54.6\%, Gaps 38.9\%, Hurdles 40.9\%, Stairs 28.8\%), others such as Steps (15.2\%) performed poorly. We suspect that policies requiring more precise foot placement benefit the most from having the target trajectory linked with the robot steps. This simple method allows us to train policies following a target, without providing the policy with a phase variable (a limitation of related work~\cite{peng_deepmimic_2018},~\cite{merel_hierarchical_2019}). Instead the policy can infer the phase of the walk cycle from joint positions and feet contact information.

\subsection{Stage 3 Analysis}
\label{sec:stage3}

% \begin{table*}[]
%     \vspace{3mm}
%     \centering
%     % \begin{tabular}{c|c|c|c|c|c}
%     \begin{tabular}{cccccc}
%         & Flat & Gap & Hurdle & Stairs & Steps \\ 
%         % \multicolumn{1}{c|}{\textbf{Stage 3 Analysis}} & Flat & Gap & Hurdle & Stairs & Steps \\ 
%         %  \multicolumn{6}{l}{\textbf{Curriculum Stage 3 Analysis}}  \\
%          \hline
%          Stage 1 and 2    & 99.8                  & \textbf{82.7} & \textbf{78.6} & 80.7                &   77.5  \\
%          Stage 1, 2 and 3 & \textbf{100.0}          & 79.3          & 70.4           & \textbf{80.9}      & \textbf{83.2}     \\

%     \hline
%     \end{tabular}
%     \caption{Stage 1 and 2, and all three stages of the curriculum, as a percentage of total distance travelled, computed with 500 trials. }
%     \label{tab:results_stage3}
% \end{table*}

\begin{figure}[tb!]
\vspace{-10mm}
\centering
\includegraphics[width=0.95\columnwidth]{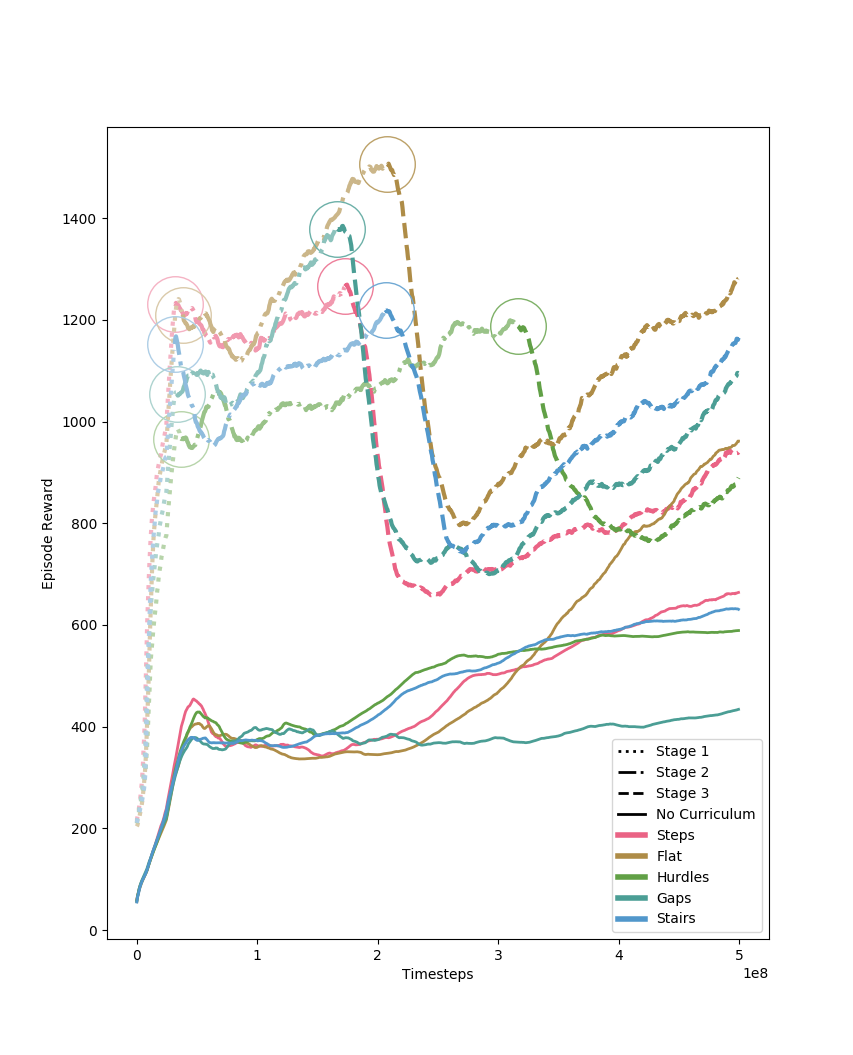}
\vspace{-10mm}
\caption{Episode rewards for our method (Stage 1, Stage 2, and Stage 3), for each terrain type. The completion of each stage is shown with a circle and change in line colour and type. The first segment (lightest colour, dotted line) is Stage 1, the middle segment shows Stage 2 (dash-dotted line), and the final segment (darkest colour, dashed line) shows the rewards when the perturbation curriculum is applied. The solid coloured lines are the policies trained without a curriculum.}
\label{fig:stage3}
\vspace{-3mm}
\end{figure}

We now analyse the final stage of our curriculum learning method. In the perturbation curriculum the maximum magnitude of external forces is increased from an initial disturbance of $p_1 = \SI{50}{\newton}$ to a final disturbance of $p_n = \SI{1000}{\newton}$, where perturbations are sampled uniformly for each degree of freedom of the CoM from $\mathcal{U}(-p, p)$. Fig~\ref{fig:stage3} shows the average episode reward for all three stages of training for all terrain types. We can see that Stage 1 completes early in training for all terrain types. Drops in reward are seen early in Stage 2 where the majority of guidance forces are removed, then we see a gradual increase as the agent learns to act without guidance. In Stage 3 we see a large drop in return as the perturbations are increased. At this point in training the agent is operating in the hardest conditions (terrain at highest difficulty, no guiding forces, large perturbations applied to the CoM). We also show the episode reward for policies trained without any part of the curriculum (solid line), and observe the reward for policies trained without a curriculum is much lower than the policies trained with our method.

From Table~\ref{tab:results_stage3} we can see that adding the perturbation curriculum improves performance slightly for all terrain types except the Hurdles terrain (58.5\% compared to 79.0\%). The Hurdles terrain takes the longest for Stage 1 and Stage 2 to complete (as shown in Fig~\ref{fig:stage3}), and therefore has the least time to train with large perturbations. 

% We note performance could be improved with perturbation tuning for individual terrain types, and that perturbations that are too large can reduce fine grained behaviour improvements.

% did not perform better on the test world than Stage 1 \& 2 (with the exception of the flat terrain: 92.4\% compared to 95.7\%). This can be explained by the Stage 1 \& 2 approach overfitting to the training domain, where increasing perturbations reduces this effect. In the below experiments we test this claim by applying perturbations during evaluation, and assess performance on terrain difficulties that are increased from what are seen during training.
%  We use a single jump from the initial disturbance level to the final disturbance level.

\begin{table}[h!]
    \vspace{-2mm}
    \centering
    \begin{adjustbox}{max width=1.0\columnwidth}  
    \begin{tabular}{cccccc}
        %  \multicolumn{1}{c|}{\textbf{Stage 1 and Stage 2 Analysis}} & Flat & Gap & Hurdle & Stairs & Steps \\
          & Flat & Gap & Hurdle & Stairs & Steps \\
         \hline
         Stage 1 \& 2    &   89.8            & 69.5           &  \textbf{79.0}          & 55.8 & 58.1  \\
         Stage 1, 2 \& 3   &  \textbf{99.9}    & \textbf{72.3}  & 58.5  & \textbf{57.6}  & \textbf{60.5}  \\
        %  Large (1000N)   &  93.0    & 59.5  & 0.0  & 18.9  & 13.2  \\
    \hline
    \end{tabular}
    \end{adjustbox}
    \caption{Stage 1 \& 2, and Stage 1,2 \& 3 evaluated as percentage of total distance travelled, computed with 500 trials, with no perturbations applied during evaluation.}
    \label{tab:results_stage3}
    \vspace{-2mm}
\end{table}

% Stage 2 completes We observed that Stage 1 usually takes the first 10-20\% of total training time; after that the terrain is at its most difficult for the remainder of training.

% \subsubsection{Effect of Stage 3}
% We compare the results of including Stage 3 in Table~\ref{tab:results_stage3}, where the first row shows the results for evaluations with Stage 1 and Stage 2, and the second row includes Stage 3. We can see that adding Stage 3 improves the performance of the policies for all terrains.

\subsubsection{Effect of perturbation intensity}
\label{sec:intense}
In this section we investigate the robustness of our method by applying perturbations during the evaluation. The evaluation perturbation forces are sampled uniformly from $(-\SI{1000}{\newton}, \SI{1000}{\newton})$ and applied to the CoM of the robot. We compare training with small perturbations, training with large perturbations, and our perturbation curriculum.

\begin{itemize}[leftmargin=*]

\item\textbf{Small:} Small perturbations are applied throughout training, sampled uniformly from $(\SI{-50}{\newton}, \SI{50}{\newton})$. This is the same as Stage 1 \& 2.

\item\textbf{Large:} Maximum perturbations are applied throughout training, sampled uniformly from $(\SI{-1000}{\newton}, \SI{1000}{\newton})$.

\item\textbf{Curriculum:} We show the results of the perturbation curriculum (Stage 3), with perturbations applied during evaluation.

\end{itemize}

\begin{table}[h!]
    \vspace{-2mm}
    \centering
    % \begin{tabular}{c|c|c|c|c|c}
    \begin{adjustbox}{max width=1.0\columnwidth}
    \begin{tabular}{cccccc}
         & Flat & Gap & Hurdle & Stairs & Steps \\ 
        %  \multicolumn{6}{l}{\textbf{Curriculum Stage 3 Analysis}}  \\
         \hline
         Small (50N) & 45.9          & 17.8          & 28.5           & 31.6          & 21.7     \\
         Large (1000N) & 81.2          & 18.0          &  21.3           & 22.7          &  9.4   \\
         Curriculum (50N to 1000N)  & \textbf{87.1}          & \textbf{35.7}          & \textbf{28.6}  & \textbf{45.6} & \textbf{24.5}   \\
    \hline
    \end{tabular}
    \end{adjustbox}
    \caption{Effect of perturbations during training, shown as a percentage of total distance travelled, computed with 500 trials. Each method is evaluated with maximal perturbations.}
    \label{tab:results_perturbations}
    \vspace{-2mm}
\end{table}

From Table~\ref{tab:results_perturbations}, we can see that our perturbation curriculum (Stage 3) performs best on all terrains compared with applying small (Stage 1 and Stage 2), and large forces throughout training. We notice slight performance gains for Hurdles (28.5\% compared to 28.6\%) and Steps (21.7\% compared to 24.5\%), and larger improvements with Flat (45.9\% compared to 87.1\%), Gaps (17.8\% compared to 35.7\%), and Stairs (31.6\% compared to 45.6\%). For policies trained with large disturbances throughout training we observe reduced performance compared with our method. Our results show that a perturbation curriculum allows for large perturbations during training, with improved outcomes compared to large forces applied for the entirety of training. This methodology can be applied to scenarios where training with perturbations can improve robustness (e.g. simulation to real transfer).

% and even failure to learni compared with either worse that small perturbations for some terrain types(, Hurdles, Stairs) and detrimental for Steps 9.4\%. 

% We show that using the perturbation curriculum improves robustness (for all terrain types), compared to applying small perturbations throughout training.  

\subsubsection{Effect of test domain difficulty}
\label{sec:diff}

We test the generalisation of our method (Stages 1, 2 \& 3) in a final evaluation that compares policy performance on terrains that are less difficult, and terrains that are more difficult than the final terrain experienced during training.

\begin{itemize}[leftmargin=*]

\item\textbf{Easy Difficulty:} We evaluate our policies with the easy terrain difficulty $d_1$.

\item\textbf{Reduced Difficulty:} We evaluate our policies with a terrain difficulty 20\% lower than the final difficulty experienced in training ($d_8$). 

\item\textbf{Normal Difficulty:} We evaluate our policies with the final terrain difficulty experienced in training ($d_{10}$). 

\item\textbf{Increased Difficulty:} We evaluate our policies with a terrain difficulty 20\% higher than the final difficulty experienced in training ($d_{12}$). 
 
\end{itemize}

\begin{table}[h!]
    \vspace{-2mm}
    \centering
    % \begin{tabular}{c|c|c|c|c|c}
    \begin{adjustbox}{max width=1.0\columnwidth}
    \begin{tabular}{cccccc}
        % \multicolumn{1}{l}{\textbf{Less Difficult}} & Flat & Gap & Hurdle & Stairs & Steps \\ 
         & Flat & Gap & Hurdle & Stairs & Steps \\              
         \hline

        Easy Difficulty           & \textbf{100.0} & 45.3            & 24.5       & \textbf{88.1}          & 12.5   \\
        Reduced Difficulty      & \textbf{100.0} & 15.4            & \textbf{62.3}       & 80.2          & \textbf{87.4}   \\
        Normal Difficulty          & 99.9    & \textbf{72.3}  & 58.5  & 57.6  & 60.5   \\
        Increased Difficulty         & 94.5          & 11.5          & 32.5   & 50.5          &  26.7   \\
    \hline
    \end{tabular}
    \end{adjustbox}
    \caption{The effect of evaluating with terrain difficulty levels that are reduced and increased from the final terrain difficulty reached during training. Shown as a percentage of total distance travelled, computed with 500 trials.}
    \label{tab:results_difficulty}
    \vspace{-2mm}
\end{table}

% From Table~\ref{tab:results_difficulty}, we can see the Gaps policy has overfit to the final difficulty experienced during training, and has failed to generalise to terrain of reduced (15.4\%) or increased (11.5\%) difficulty. For all other terrain types we observe improved performance on the reduced difficulty terrain, and comparable performance for some terrain types with increased difficulty (Flat 94.5\% compared to 99.9\%, Stairs 50.5\% compared to 57.6\%). Our method requires more investigation to improve the generalisation of our policies to terrain difficulties not seen during training.

From Table~\ref{tab:results_difficulty}, we can see how policies adapt to the terrain during training, with some policies (Gaps, Hurdles, Steps) performing poorly when evaluated with an easy terrain difficulty. The Gaps policy has overfit to the final difficulty, performing poorly on reduced (15.4\%) or increased (11.5\%) difficulty. For all other terrain types we can see improved performance on the reduced difficulty terrain, and comparable performance for some terrain types with increased difficulty (Flat 94.5\% compared to 99.9\%, Stairs 50.5\% compared to 57.6\%), suggesting generalisation in these cases. An alternative training strategy that involves training with terrains of multiple difficulties may improve policy performance over a wider range of terrain difficulties.

%%%%%%%%%%%%%%%%%%%%%%%%%%%%%%%%%%%%%%%%%%%%%%%%%%%%%%%%%%%%%%%%%%%%%%%%%%%%%%%%
\section{Conclusion}
\label{sec:conclusion}
We demonstrate a curriculum learning approach for developing DRL policies for a torque controlled biped traversing complex terrain artifacts. Our agent was successfully able to traverse several difficult terrain types where removal of any component of our multi-stage curriculum resulted in decreased performance or failure to learn. A key idea of our method is that a simple target policy is sufficient to train a policy on various terrains, where much of the behaviour is acquired from the terrain itself.

A limitation of our work is that the curriculum is manually tuned, whereas an adaptive curriculum may be able to learn harder tasks, which is left for future work. We will also investigate methods for switching between these policies such that we can expand the capabilities of our platform.

\bibliographystyle{named}
\bibliography{references}

\end{document}